%% file: main.tex
\pdfoutput = 1
\documentclass[twocolumn]{article}
\input{macros}

\usepackage{style}

\makeatletter
\@namedef{ver@everyshi.sty}{}
\newcommand{\removelatexerror}{\let\@latex@error\@gobble}

\newcommand\cl{\texttt{CL}\xspace}
\newcommand\camcos{\texttt{CAM-COS}\xspace}  
\newcommand\attcos{\texttt{ATT-COS}\xspace}  
\newcommand\aus{\texttt{AUs}\xspace}
\newcommand\au{\texttt{AU}\xspace}
\newcommand\rafdb{\texttt{RAF-DB}\xspace}
\newcommand\affectnet{\texttt{AffectNet}\xspace}

\newcommand{\normx}[1]{\ensuremath \lVert#1\rVert}

\title{Guided Interpretable Facial Expression Recognition via Spatial Action Unit Cues }

\renewcommand\footnotemark{}

\author{Soufiane~Belharbi$^{1}$,
  ~Marco~Pedersoli$^{1}$,
  ~Alessandro~Lameiras~Koerich$^{1}$,
  ~Simon~Bacon$^{2}$, and
  ~Eric~Granger$^{1}$\\
 	$^1$ LIVIA, Dept. of Systems Engineering, ÉTS, Montreal, Canada \\
	$^2$ Dept. of Health, Kinesiology \& Applied Physiology, Concordia University, Montreal, Canada\\
{\tt\footnotesize \textcolor{black}{soufiane.belharbi@etsmtl.ca} }
}

\newcommand{\ignore}[1]{}

\begin{document}
\maketitle\thispagestyle{fancy}

\maketitle
\rhead{\color{gray} \small Belharbi et al. \;  [FG 2024]}

\begin{abstract}
Although state-of-the-art classifiers for facial expression recognition (FER) can achieve a high level of accuracy, they lack interpretability, an important feature for end-users. 
Experts typically associate spatial action units (\aus) from a codebook to facial regions for the visual interpretation of expressions. 
In this paper, the same expert steps are followed. A new learning strategy is proposed to explicitly incorporate \au cues into classifier training, allowing to train deep interpretable models. 
During training, this \au codebook is used, along with the input image expression label, and facial landmarks, to construct a \au heatmap that indicates the most discriminative image regions of interest w.r.t the facial expression. This valuable spatial cue is leveraged to train a deep interpretable classifier for FER.
This is achieved by constraining the spatial layer features of a classifier to be correlated with \au heatmaps. Using a composite loss, the classifier is trained to correctly classify an image while yielding interpretable visual layer-wise attention correlated with \au maps, simulating the expert decision process. Our strategy only relies on image class expression for supervision, without additional manual annotations. Our new strategy is generic, and can be applied to any deep CNN- or transformer-based classifier without requiring any architectural change or significant additional training time.
Our extensive evaluation\footnote{Code :\href{https://github.com/sbelharbi/interpretable-fer-aus}{https://github.com/sbelharbi/interpretable-fer-aus}.} on two public benchmarks \rafdb, and \affectnet datasets shows that our proposed strategy can improve layer-wise interpretability without degrading classification performance. In addition, we explore a common type of interpretable classifiers that rely on class activation mapping (CAM) methods, and show that our approach can also improve CAM interpretability.
\end{abstract}


%
%

\begin{figure}[ht!]
\centering
  \centering
  \includegraphics[width=\linewidth]{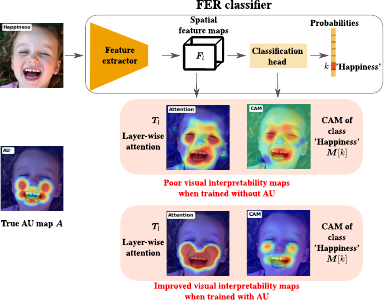}
  \caption[Caption]{Comparison of class activation mapping (CAM) and attention maps produced at inference time using a FER classifier trained without (top) and with (bottom, ours) AU maps. In our experiments, we use an identical architecture to compare the impact of training with and without \au maps. The difference only resides in an additional AU-based training loss using \au maps. 
  Training a FER classifier with our AU maps yields attention and CAM that are aligned with the expert's knowledge used to assess basic facial expressions in images~\cite{Martinez19}, as illustrated in the AU map ${\bm{A}}$. Consequently, our approach allows training a classifier that provides reliable interpretability, without compromising the classification accuracy. Details of our training strategy are presented in Fig.\ref{fig:proposal}. Note that in CAM-based models, the classification head can be fully convolutional or standard fully connected layers pooling posterior probabilities. 
  }
  \label{fig:promo}
\end{figure}

\section{Introduction}
\label{sec:intro}
Facial expression recognition (FER) has recently received much interest in the computer vision and machine learning communities~\cite{BonnardDDB22,xue2022vision, ZhengM023}. Since facial expression is one of the most important ways for people to express emotions~\cite{dawring1988}, FER finds a wide range of applications in, e.g., medical analysis and monitoring~\cite{Altameem20,kim20,yolcu2017}, e-health~\cite{ter21}, driver fatigue detection~\cite{Assari11,liu20}, safe driving~\cite{jeong2018}, security~\cite{li2021,nan22}, lecturing~\cite{tongucc20}, and many other fields~\cite{Sajjad23}.

FER remains challenging, particularly for real-world applications. This is due to the subtle differences between expressions, leading to low inter-class variability. Learning to distinguish samples from overlapping classes can be difficult. 
Additionally, people express the same facial expression differently depending on attributes such as level of expressiveness, age, gender, and ethnic background~\cite{li20survey,valstar12}. To tackle these challenges, many deep learning methods have been proposed to train highly accurate classification models~\cite{BonnardDDB22,li20survey,xue2022vision, ZhengM023}. This is achieved by learning robust feature representations compared to traditional hand-crafted features~\cite{DalalT05, Lowe04, Shan09}.
Deep FER methods usually rely on either global or local approaches to learn feature representations. Global approaches~\cite{Farzaneh21,li17} propose training loss functions to improve the overall representation robustness. To avoid incorporating noise into the global image representation, local approaches resort to learning from parts of the image~\cite{BonnardDDB22,happy14,li18,wang20,xie18,xie19deep,xue2022vision, ZhengM023}. These methods use facial landmarks to extract robust local features around relevant facial parts~\cite{happy14,xie18, ZhengM023}. Other methods rely on self-attention mechanisms to focus on relevant and discriminative parts of the facial image and suppress noisy regions~\cite{BonnardDDB22,li18,wang20,xie19deep,xue2022vision}.

State-of-the-art FER classifiers have achieved significant progress in terms of accuracy. However, end-users may not require only an accurate classifier that yields a classification score in multiple critical applications~\cite{deramgozin22,puente19,zhou18}. They may also need the model to provide an interpretable decision~\cite{deramgozin22,yadav22,zhou18}. For instance, this can help clinicians and therapists understand and build trust in the FER model decisions, planning better future interventions~\cite{vellido20}. Subsequently, this facilitates better integration of machine learning methods into daily clinical routines and health care practice~\cite{vellido20}. In addition, interpretability can greatly help diagnose errors made by machine learning models and facilitate the identification of weaknesses for future improvement. Unfortunately, interpretability has been overlooked in the FER due to the main focus on classification accuracy. This has led to the design of FER models that lack interpretability. Recent progress with attention-based methods like  APVIT~\cite{xue2022vision} allows extracting internal attention explicitly to select only discriminative regions of interest (ROIs). Although they provide visual interpretability to highlight the regions used for classification, such ROIs are not necessarily aligned with expert knowledge typically used to assess facial expressions.

\begin{figure*}[ht!]
\centering
  \centering
  \includegraphics[width=0.9\linewidth]{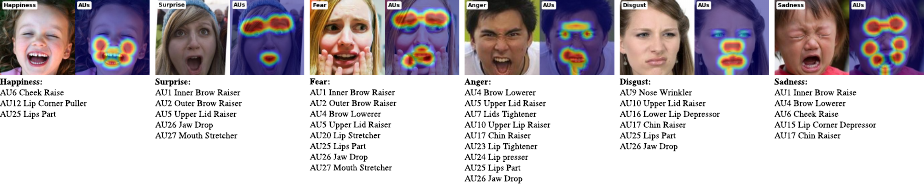}
  \caption[Caption]{Codebook of basic facial expressions and their associated \aus~\cite{Martinez19}. The spatial \au map is built using the image expression to select the right corresponding \au subset in combination with facial landmarks, which are employed to localize these same \aus in the image. In particular, the location of landmarks is used to estimate \au positions. For instance, the right 'Cheek' location is estimated using landmark 47 (middle of the low right eye) and 11 (right side of the jaw).
  The code 'AUx' is the identifier of the \au~\cite{Martinez19}. }
  \label{fig:codebook-fer-aus}
\end{figure*}

Experts commonly rely on a codebook of AUs to determine a basic facial expression~\cite{Martinez19} (Fig.~\ref{fig:codebook-fer-aus}). Each expression is associated with a subset of spatial AUs in the face. Ideally, an interpretable classifier should point to and localize the correct AU ROIs when predicting the corresponding expression. However, this task is challenging since predicting such ROIs using only image expression supervision is not trivial. Additionally, the localization cues are not publicly available to be learned due to the annotation costs and the complexity of building large-scale benchmarks.

This paper introduces a generic learning strategy to build an interpretable deep classifier for the FER task. This is achieved using spatial AU cues constructed from the image class supervision without extra manual annotation cost. This explicitly integrates the expert facial expression assessment approach into a classifier's decision-making. In particular, we build a visual interpretability tool consistent with \au cues used by experts~\cite{Martinez19}. To this end, a spatial discriminative heatmap is constructed that gathers relevant locations of \aus required to determine the expression in an input image. Then, layer-wise deep features of a classifier are constrained to be correlated with such heatmap cues during training. At the same time, the model is trained to classify the image correctly. Such a multi-task training scheme allows for building accurate but, most importantly, interpretable classifiers. During inference, the layer-wise attention provides a visual interpretability tool that indicates which ROIs are used to determine the predicted expression (Fig.~\ref{fig:proposal}, \ref{fig:promo}). This is achieved without additional annotation costs, as illustrated in Fig.~\ref{fig:proposal}. A discriminative AU heatmap can be built simply using image class expression, an AU codebook, and facial landmarks, which are commonly estimated using off-the-shelf methods. 
Interpretability in models usually emerges as a result of other tasks, such as classification without explicit supervision for interpretability~\cite{doshi2017}. It represents a tool to clarify the model decision, for instance, under the form of visual ROIs. However, since we explicitly provide interpretability cues to train our model, we refer to it as \emph{guided} interpretable FER model.

Following the interpretable classifier direction, we explore a curated branch of classifiers in this work that allows us to perform image classification and yield visual interpretability. In particular, we employ CAM-based classifiers~\cite{choe2020evaluating,rony2023deep}. These methods are popular for weakly-supervised object localization (WSOL) tasks in computer vision on different imaging types, including natural scene~\cite{zhou2016learning} and medical imaging~\cite{rony2023deep,wang24}. Given an input image, a classifier can be trained using only image-class labels to correctly classify an image and provide a per-class heatmap, i.e., CAM, to localize image ROIs related to the output prediction. Therefore, they play an important role in localization and visual interpretability, making them well-suited for our work. Consequently, we conduct a comparative study using different CAM methods from other families of methods, and assess their capacity for classification and interpretability on the FER task. Note that this is the first work to leverage CAM methods for the FER task.

\noindent \textbf{Our main contributions are summarized as follows:}

\noindent \textbf{(1)} To improve the visual interpretability of state-of-the-art FER classifiers, we introduce a learning strategy allowing to training of an accurate but, most importantly, interpretable deep classifier (Fig.\ref{fig:promo}) that is consistent with the process used by experts to determine basic facial expressions. Our training relies on aligning spatial features with spatial \au maps built using facial landmarks and image-class labels. This guides the classifier's attention to use regions around \aus leading to more interpretable decisions.

\noindent \textbf{(2)} Our method does \emph{not} require extra manual annotation, significant extra computations during training, nor change the model architecture or the inference process. In addition, our method is generic -- it can be used with any deep CNN or transformer-based model.

\noindent \textbf{(3)} Our approach is validated experimentally on two public FER benchmarks (\rafdb, and \affectnet) in terms of classification and interpretability accuracy. We evaluated different CAM-based methods with and without our spatial \au cues. Different ablations are provided as well. Empirical results showed that both classification and interpretability improved. Interestingly, compared to a vanilla deep classifier, we show that its layer-wise attention can largely be shifted using our \au spatial cues without compromising classification accuracy. We show that classification accuracy improves, particularly over large datasets such as \affectnet. This demonstrates that spatial \aus are a reliable source for discriminative ROIs for basic facial expression recognition~\cite{Martinez19}.

\section{Related Work}
\label{sec:related-w}
This review covers related work on FER classifier systems, CAM methods for localization and interpretability, FER systems interpretability, and \aus in FER tasks.

\subsection{Standard FER classifiers} 
Given the availability of large public benchmarks~\cite{li17,MollahosseiniHM19}, different methods have been proposed to achieve state-of-the-art classification accuracy~\cite{li20survey}. This is most achieved by designing robust features to overcome the limitations of traditional hand-crafted features~\cite{DalalT05, Lowe04, Shan09}. Robust features are also learned using deep models, either through a global or local approach. 

Global methods typically focus on designing robust training losses to build enhanced discriminative features while using the entire image to build a global representation~\cite{Farzaneh21,li17}. For instance, Farzaneh and Qi~\cite{Farzaneh21} leverage deep metric learning and propose a deep attentive center loss to select a subset of relevant features for classification adaptively. Although successful, image faces are typically noisy, which may easily corrupt the features. Local methods tackle this issue by explicitly incorporating different mechanisms to remove the noise and build better features. Mainly, these methods rely on a part-based approach assuming that only some regions in the image are relevant to determine the expression~\cite{BonnardDDB22,happy14,li18,wang20,xie18,xie19deep,xue21,xue2022vision, ZhengM023}. 

A subset of these methods assumes that discriminative features are located around facial landmarks. Therefore, only these regions are cropped either at image or spatial feature level~\cite{happy14,xie18, ZhengM023}. This assumes that relevant regions are located around the landmarks. Such early dropout of patches may discard relevant patches not included in the landmarks in addition to missing the context. This also requires highly accurate landmarks. However, leveraging landmarks has been successful in the recent work POSTER~\cite{Mao23, ZhengM023} by cross-fusion of sparse key-landmarks features with standard global image features.

A second subset exploits attention mechanism~\cite{BonnardDDB22,xie19deep,wang20,xue21,xue2022vision} either self-learned or with provided cue.
DAM-CNN~\cite{xie19deep} learns a spatial feature attention to filter out noise and keep only relevant spatial features to build a dense embedding later.
The APVIT method~\cite{xue21,xue2022vision} leverages transformers and their attention potential to design a FER classifier. In particular, they propose a self-attention approach early in the network. Such attention allows the selection of relevant patch regions and performs hard attention at the spatial feature level. Patches are scored, and only the top-k are allowed to proceed into the next layer to build an image embedding. This explicitly allows the model to learn to filter out irrelevant patches.
Other methods are provided external spatial cues to be aware \emph{where} to look for discriminative regions~\cite{BonnardDDB22,li18,wang20}. 
RAN~\cite{wang20} is a region-based method. It performs image cropping either randomly or using landmarks. Their features are then attended using self-attention and relation-attention modules. Such attention is then used to combine different features and build an image embedding. This makes the final representation robust to pose and occlusion. Li et al.~\cite{li18} follow a similar direction by decomposing the spatial features into the regions and using the attention module afterward.
Bonnard et al.~\cite{BonnardDDB22} use facial landmarks to build a heatmap, which is used to align spatial features in a deep model. This is expected to filter out irrelevant spatial features and focus mostly on features around landmarks. Our work is more aligned with this family of methods. However, we use grounded and more reliable cues, that is, spatial \aus. Since spatial \aus are class-dependent, they are more discriminative for the FER task~\cite{Martinez19}, compared to facial landmarks, which are generic and class-agnostic.

While the abovementioned methods focus on FER over still images, other methods tackle video applications~\cite{liu23facial,liu2021video,liu21identity,liu23}. These methods deal with similar issues presented in still images while leveraging temporal dependency and multi-modality in videos to capture better expressiveness and build robust features.

\subsection{CAM methods for WSOL and interpretability}
Using only image class as supervision, CAM-based methods can be trained to classify an image correctly and yield reliable localization of ROIs related to the image class. 
They are currently the dominant approach for the WSOL task~\cite{choe2020evaluating,rony2023deep}. They achieved interesting results in different domains, including natural scene images~\cite{belharbi2022fcam,rony2023deep, Wu23}, and medical imaging~\cite{negevsbelharbi2022,belharbi2020minmaxuncer,rony2023deep,wang24}. They have also been extended for localization in videos~\cite{tcamsbelharbi2023,belharbi2023colocam}.
Early works have focused on building different spatial pooling layers~\cite{durand2017wildcat,durand2016weldon, oquab2015object,zhou2016learning}.  
However, CAMs tend to highlight only small and most discriminative parts of an object~\cite{choe2020evaluating,rony2023deep}, limiting its localization performance. To overcome this, different strategies have been proposed, including data augmentation on input image or deep features~\cite{belharbi2020minmaxuncer, ChoeS19, MaiYL20eil}, as well as architectural changes~\cite{gao2021tscam, LeeKLLY19, Wu23, ZhangW020i2c}. 

Recently, learning via pseudo-labels has shown great potential, despite its reliance on noisy labels~\cite{belharbi2022fcam,negevsbelharbi2022, MeethalPBG20icprcstn, Murtaza2023dips,murtaza2022dipssypo,murtaza2022dips,wei2021shallowspol}. Most previous methods used only forward information in a CNN (bottom-up family~\cite{rony2023deep}). However, other methods also aim to leverage backward information (top-down methods~\cite{rony2023deep}). These include biologically inspired methods~\cite{cao2015look,zhang2018top}, and gradient~\cite{ChattopadhyaySH18wacvgradcampp,fu2020axiom,JiangZHCW21layercam,SelvarajuCDVPB17iccvgradcam} or confidence score aggregation methods~\cite{desai2020ablation,naidu2020iscam}. CAM-based methods have been used for interpretability as well~\cite{Samek2019,xiao23} since they provide a map that indicates ROIs relevant to the model's decision. Gradient-based CAMs are most common in interpretability task~\cite{ChattopadhyaySH18wacvgradcampp,fu2020axiom,JiangZHCW21layercam,SelvarajuCDVPB17iccvgradcam} in addition to~\cite{cao2015look,zhang2018top}. They mainly search for ROIs in the \emph{feature maps} that better stimulate a class response. This inspires other methods~\cite{zhang21} to leverage gradient and extend it to input image and perform perturbation analysis~\cite{dabkowski17,fong19,fong17, Petsiuk18, ZeilerF14} where the aim is to find which ROI of the network's input that better stimulates its output. Compared to CAM-based, these methods are often more expensive in computation and require optimization, making them less attractive for FER tasks. However, CAM-based methods are simple to use, and they are built into the model, requiring a single forward or a forward and backward computation.

\subsection{Interpretability in FER systems} 
In critical FER applications such as e-health and behavioral health interventions and assessments~\cite{chen22}, it may not be enough to predict the facial expression accurately. Users may require interpretability to help understand the model decision~\cite{deramgozin22,puente19,yadav22,zhou18}.
Although important, interpretability in FER systems has been largely overlooked. This is mainly due to the focus on achieving state-of-the-art classification accuracy over challenging benchmarks~\cite{li20survey}. The absence of public datasets with interpretability annotation has also contributed to shadowing such an important task. Some recent works have attempted to provide built-in discriminative attention under visual interpretability, such as the APVIT method~\cite{xue21,xue2022vision}. Due to the lack of annotation, these methods are still limited and unreliable since their interpretability has not yet been quantified.
In this work, we propose a protocol to evaluate the interpretability of a FER system without the need for extra manual annotation. We use spatial facial \aus to simulate experts' processes to assess expressions~\cite{Martinez19}. Such \aus are built and encoded automatically in a convenient spatial heatmap using only image expression as supervision. This can help to automatically label large benchmarks easily, allowing training and evaluation of FER models in terms of interpretability.

\subsection{\aus in FER systems} 
\begin{figure}[ht!]
\centering
  \centering
  \includegraphics[width=\linewidth]{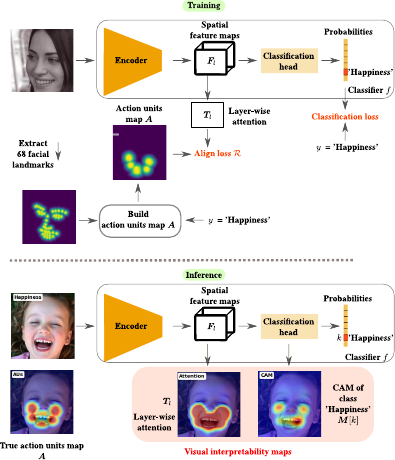}
  \caption[Caption]{Our interpretable classifier for the FER task (training and inference). Each basic facial expression can be determined via a set of \aus~\cite{Martinez19}. Therefore, to train our interpretable FER classifier, we first extract facial landmarks and build a \emph{discriminative} spatial map ${\bm{A}}$ that contains the set of all \aus associated with the image class expression~\cite{Martinez19}. This map is used as localization cues to train layer-wise attention ${\bm{T}_l}$ to focus on the ROIs highlighted in the \au map. A classification loss, such as cross-entropy, is also used. Once trained, the classifier yields an interpretable layer-wise attention map. When a CAM method~\cite{rony2023deep} is considered, the classifier can also produce a per-class interpretable map.}
  \label{fig:proposal}
\end{figure}

The Facial Action Coding System (FACS) is a taxonomy for fine-grained facial expression analysis~\cite{ekman1978,friesen1978}. They have long been used to analyze facial expressions~\cite{fan20,jacob21, Luo22, Sanchez18,zhang18bilat}. Each basic facial expression is associated with a list of \aus as they determine which facial muscles are involved in expressing such emotion. The standard established task in the literature is \au detection~\cite{jacob21, Luo22}. It is a supervised multi-label classification task. It aims at predicting the correct set of \emph{active} \aus in an input image. To perform such a task, a tedious annotation is required to determine the \emph{active} set of \aus in each image since not all \aus must be active at once. Another related task goes further to estimate the \emph{intensity} of the \aus~\cite{fan20,zhang18bilat}, which is furthermore challenging. Other works aim to localize and estimate the intensity~\cite{Sanchez18} jointly. 
The work of~\cite{jacob21} is relatively close to our work. To accurately detect active \aus through multi-label classification, authors leverage multi-task learning to predict a localization heatmap for the same \aus jointly. The goal is to improve the detection of \aus via their spatial localization. However, \aus have not been used for the interpretability of FER classifiers. This makes our work the first to do so. To avoid the extra annotation costs, we used a generic discriminative \aus codebook defined in~\cite{Martinez19}. It allows the automatic labeling of extensive benchmarks without manual intervention (Fig.~\ref{fig:proposal}). The codebook associates a set of \aus to a basic facial expression (Fig.~\ref{fig:codebook-fer-aus}). This is used to build a discriminative heatmap that holds potential ROIs to be inspected by the model to determine the facial expression in the input image. This does not require all the \aus to be active. But, it is more likely that part of them are active for the expression to manifest in an image. A further step of our work is to use active \aus to build an ideal interpretable model, although this comes at an additional expensive annotation cost. Alternatively, a pretrained \au detector, such as~\cite{jacob21}, could be used.
 
\section{Proposed Approach}
\label{sec:method}

\subsection{Notation} 
We denote by ${\mathbb{D} = \{(\bm{X}, y)_i\}_{i=1}^N}$ a training set of $N$ images, where ${\bm{X}_i}$, of size ${h\times w}$, is the input image, and ${y_i \in \{1, \cdots, K\}}$ is its image class, i.e. facial expression, with $K$ is the total number of classes. We denote by ${f(\bm{X}; \bm{\theta})}$ a deep classifier with ${\bm{\theta}}$ its parameters. The posterior per-class probabilities are referred to as ${f(\bm{X}) \in [0, 1]^K}$ where ${f(\bm{X})_k}$ is the probability of the class ${k}$. 
The spatial feature maps produced by the classifier's encoder at a layer ${l}$ are a tensor denoted by ${\bm{F}_l \in \reals^{n\times h^{\prime}\times w^{\prime}}}$, where $n$ is the number of feature maps, and ${h^{\prime}\times w^{\prime}}$ is their size which is typically smaller than the input image. 
CAM-based WSOL methods produce an additional spatial tensor that holds all the CAMs, ${\bm{M} \in \reals^{K\times h^{\prime\prime}\times w^{\prime\prime}}}$, where ${\bm{M}[k]}$ is the CAM of the class ${k}$.

\subsection{Building spatial \au maps} 
Given an input image and its label ${(\bm{X}, y)}$, standard 68 facial landmarks are extracted from the image using off-the-shelf techniques such as SynergyNet~\cite{wu21}. This allows us to \emph{localize} relevant parts of the face such as mouth, nose, eyes, and eyebrows, which will be helpful later to estimate the location of relevant \aus. Martinez et al.~\cite{Martinez19} suggest that each facial expression can be recognized by inspecting a combination of \aus (Fig.~\ref{fig:codebook-fer-aus}). For instance, the expression 'Happiness' involves a set of \aus: 'cheek raise', 'lip corner puller', 'lips part'. Using the extracted facial landmarks around the mouth part, we can localize the 'lip part' of the face. Such localization is translated into a 2D heatmap where strong activations are concentrated around the 'lip' to indicate an ROI and tiny activations everywhere else to indicate the background. Such a 2D map is of great importance in training since it provides useful localization cues for relevant discriminative parts in an input image for the FER task used by experts. Typically, this requires manual annotation to delineate ROIs, increasing the annotation cost. Unfortunately, such valuable annotation is not available in public facial expression benchmarks. This work simulates such ROIs \emph{without} an additional manual supervision cost.
Using the codebook of facial expressions and their associated \aus provided in~\cite{Martinez19} (Fig.~\ref{fig:codebook-fer-aus}), in combination with the extracted facial landmarks and the image label expression ${y}$, a single 2D heatmap, ${\bm{A} \in \reals^{h\times w}}$, is built to hold all the relevant (i.e., discriminative) \aus for the input image (Fig.~\ref{fig:proposal}). Consequently, the input image is augmented with an extra supervision cue to be ${(\bm{X}, y, \bm{A})}$ where ${\bm{A}}$ is the estimated \au map. To ease subsequent alignment computations, this map is normalized to have values between 0 and 1: ${\bm{A} \in [0, 1]^{h\times w}}$. Note that this process requires only the image class, i.e., image facial expression, as supervision. No extra manual annotations are needed.

\subsection{Layer-wise attention}
Given a layer-wise spatial feature tensor ${\bm{F}_l}$ at layer $l$, we aim to determine which spatial parts are relevant for classification. A common approach to achieve this is through features self-attention~\cite{ChoeS19,zhu2017soft}. Such spatial attention is estimated via the average feature map as follows,
\begin{equation}
    \label{eq:attention}
    \bm{T}_l = \frac{1}{n}\sum_{j = 0}^n \bm{F}_l[j] \;,
\end{equation}
where $n$ is the total number of feature maps. The discrepancy in activations in ${\bm{T}_l}$ differentiates relevant from irrelevant regions. High activations indicate potential spatial ROIs for classification, while low activations point to background and noisy regions. In a deep classifier, such attention maps are a great candidate for incorporating spatial learning cues into the model. Although CAM tensors ${\bm{M}}$ hold spatial discriminative cues, they are sensitive to being altered. Aligning them explicitly with other maps can easily lead to poor classification~\cite{belharbi2022fcam,choe2020evaluating}. Therefore, we rely on layer-wise attention to learn better and interpretable spatial features.

\subsection{\au map alignment loss}
After estimating both layer-wise attention ${\bm{T}_l}$ and \au map ${\bm{A}}$, we can proceed with their alignment to train the classifier to yield similar spatial cues. Since the dynamic range of the attention map is unknown beforehand, and since it changes during training, it is inadequate to train the attention to have the same values as the \au map ${\bm{A}}$. Instead, we resort to a loss that aims to yield attention maps that are \emph{correlated} with ${\bm{A}}$. To this end, we use cosine similarity as,
\begin{equation}
    \label{eq:cosine}
    \mathcal{R}(\bm{T}_l, \bm{A}) = \frac{\sum \left(\bm{T}_l \odot \bm{A} \right) }{\normx{\bm{T}_l}_2 \normx{\bm{A}}_2} \;,
\end{equation}
where ${\cdot \odot \cdot}$ is the Hadamard product, and ${\normx{\cdot}_2}$ is the $\ell^2$ norm. Maximizing Eq.~\eqref{eq:cosine} pushes the layer $l$ to learn spatial features in a way their mean is highly correlated with the \au map cue ${\bm{A}}$. In practice, the map ${\bm{A}}$ is downsampled to the same size as ${\bm{T}_l}$ before applying the alignment in Eq.~\eqref{eq:cosine}.

\subsection{Total training loss}
Given the triplet ${(\bm{X}, y, \bm{A})}$, our goal is to train the classifier ${f}$ on input image ${\bm{X}}$ to yield the correct class ${y}$. Additionally, we aim to encourage layer ${l}$ to construct spatial features that are correlated with the \au map cues ${\bm{A}}$. To achieve this, we jointly minimize the following composite loss,
\begin{equation}
    \label{eq:total-loss}
    \min_{\bm{\theta}} \quad - \log(f(\bm{X};\bm{\theta})_y) + \lambda (1 - \mathcal{R}(\bm{T}_l, \bm{A}))\;,
\end{equation}
where ${\lambda \geq 0}$ is a weighting coefficient that balances the importance of attention alignment with the \au map compared to cross-entropy loss (left side term). The generalization of Eq.~\eqref{eq:total-loss} to a combination of multiple layers is straightforward. It can be achieved by adding more layer-wise terms to the loss. Minimizing Eq.~\eqref{eq:total-loss} incorporates explicitly the experts' procedure in assessing basic facial expressions in images into the model decision process. The classifier is trained to localize the relevant \aus to build discriminative features for expression prediction. This justifies using layer-wise attention ${\bm{T}_l}$ as an interpretability tool during inference time (Fig.~\ref{fig:proposal}).

We note that the computational cost added by the alignment term is negligible since it can be easily computed on GPU. This leads to a training time similar to the vanilla case, where no alignment is used. The only required pre-processing is facial landmark extraction, which can be done offline once and stored on disk. \au map ${\bm{A}}$ can be computed in a negligible time on CPU on the fly during training using the image-class label only as supervision.
Since image-class labels are unavailable at test time, \au maps can not be built using the lookup table emotion-\aus. Hence, training the model to produce them along with the true label is more realistic and practical. Furthermore, it allows the model itself to build them, leading to more interpretable predictions.

{
\setlength{\tabcolsep}{3pt}
\renewcommand{\arraystretch}{1.1}
\begin{table*}[ht!]
\centering
\resizebox{.7\textwidth}{!}{%
\centering
\small
\begin{tabular}{|l|c*{3}{c}c|c*{3}{c}c|}
\cline{1-11}
&& \multicolumn{4}{c}{\rafdb} && \multicolumn{4}{c|}{\affectnet} \\
\cline{2-11}
&& \multicolumn{2}{c}{\cl} & \multicolumn{2}{c}{\camcos} && \multicolumn{2}{c}{\cl} & \multicolumn{2}{c|}{\camcos} \\
\cline{2-11}
\textbf{Method}  && w/o \au & w/ \au &  w/o \au & w/ \au &&  w/o \au & w/ \au &  w/o \au & w/ \au \\
\cline{1-11} 
\textbf{CNN-based}  &  \multicolumn{10}{c|}{} \\
CAM~\cite{zhou2016learning} {\small \emph{(cvpr,2016)}}                     && $88.20$  & $\bm{88.95}$  & $0.55$  & $\bm{0.70}$  && $60.88$ & $\bm{62.37}$ & $0.56$  & $\bm{0.69}$ \\
WILDCAT~\cite{durand2017wildcat} {\small \emph{(cvpr,2017)}}                && $88.26$  & $\bm{88.85}$  & $0.52$  & $\bm{0.69}$  && $59.88$ & $\bm{61.62}$ & $0.62$  & $\bm{0.80}$ \\
GradCAM~\cite{SelvarajuCDVPB17iccvgradcam} {\small \emph{(iccv,2017)}}      && $88.39$  & $\bm{88.85}$  & $0.55$  & $\bm{0.74}$  && $60.77$ & $\bm{62.08}$ & $0.53$  & $\bm{0.75}$ \\
GradCAM++~\cite{ChattopadhyaySH18wacvgradcampp} {\small \emph{(wacv,2018)}} && $87.84$  & $\bm{89.14}$  & $0.60$  & $\bm{0.82}$  && $60.22$ & $\bm{62.45}$ & $0.66$  & $\bm{0.83}$ \\
ACoL~\cite{ZhangWF0H18} {\small \emph{(cvpr,2018)}}                         && $87.94$  & $\bm{88.68}$  & $0.54$  & $\bm{0.67}$  && $58.28$ & $\bm{61.48}$ & $0.55$  & $\bm{0.65}$ \\
PRM~\cite{ZhouZYQJ18PRM}  {\small \emph{(cvpr,2018)}}                       && $88.13$  & $\bm{88.88}$  & $0.48$  & $\bm{0.59}$  && $57.77$ & $\bm{60.97}$ & $0.52$  & $\bm{0.75}$ \\
ADL~\cite{ChoeS19} {\small \emph{(cvpr,2019)}}                              && $87.45$  & $\bm{88.65}$  & $0.50$  & $\bm{0.63}$  && $57.88$ & $\bm{61.25}$ & $0.54$  & $\bm{0.66}$ \\
CutMix~\cite{YunHCOYC19} {\small \emph{(eccv,2019)}}                        && $88.39$  & $\bm{88.59}$  & $0.55$  & $\bm{0.57}$  && $58.74$ & $\bm{59.88}$ & $0.56$  & $\bm{0.58}$ \\
LayerCAM~\cite{JiangZHCW21layercam} {\small \emph{(ieee,2021)}}             && $87.90$  & $\bm{88.88}$  & $0.60$  & $\bm{0.84}$  && $60.77$ & $\bm{62.45}$ & $0.66$  & $\bm{0.83}$ \\
\cline{1-11} 
\textbf{Transformer-based}  &  \multicolumn{10}{c|}{} \\
TS-CAM~\cite{gao2021tscam} {\small \emph{(iccv,2021)}}                      && $86.70$  & $\bm{88.00}$  & $0.58$  & $\bm{0.71}$  && $58.99$ & $\bm{59.54}$ & $0.57$  & $\bm{0.58}$ \\
APViT~\cite{xue2022vision} {\small \emph{(ieee,2022)}}                      && $91.00$  & $\bm{91.03}$  & $--$    & $--$         && $60.62$ & $\bm{62.28}$ & $--$    & $--$        \\
\cline{1-11} 
\end{tabular}
}
\caption{\textbf{Classification (\cl) and CAM-localization (\camcos)} performance on \rafdb and \affectnet test sets with and without \aus across methods.}
\label{tab:cl-camloc}
\end{table*}
}

{
\setlength{\tabcolsep}{3pt}
\renewcommand{\arraystretch}{1.1}
\begin{table}[ht!]
\centering
\resizebox{\linewidth}{!}{%
\centering
\small
\begin{tabular}{|l|c*{2}{c}c*{2}{c}|}
\cline{1-7}
&& \multicolumn{2}{c}{\rafdb}  && \multicolumn{2}{c|}{\affectnet}  \\
\cline{2-7} 
Methods / Case  && w/o \au & w/ \au &&  w/o \au & w/ \au \\
\cline{1-7} 
\textbf{CNN-based}  &&  \multicolumn{5}{c|}{} \\
CAM~\cite{zhou2016learning} {\small \emph{(cvpr,2016)}}                     && $0.57$  & $\bm{0.85}$  && $0.64$  & $\bm{0.82}$ \\
WILDCAT~\cite{durand2017wildcat} {\small \emph{(cvpr,2017)}}                && $0.47$  & $\bm{0.85}$  && $0.61$  & $\bm{0.81}$ \\
GradCAM~\cite{SelvarajuCDVPB17iccvgradcam} {\small \emph{(iccv,2017)}}      && $0.63$  & $\bm{0.85}$  && $0.65$  & $\bm{0.82}$ \\
GradCAM++~\cite{ChattopadhyaySH18wacvgradcampp} {\small \emph{(wacv,2018)}} && $0.52$  & $\bm{0.87}$  && $0.65$  & $\bm{0.82}$ \\
ACoL~\cite{ZhangWF0H18} {\small \emph{(cvpr,2018)}}                         && $0.46$  & $\bm{0.84}$  && $0.60$  & $\bm{0.81}$ \\
PRM~\cite{ZhouZYQJ18PRM}  {\small \emph{(cvpr,2018)}}                       && $0.43$  & $\bm{0.85}$  && $0.55$  & $\bm{0.82}$ \\
ADL~\cite{ChoeS19} {\small \emph{(cvpr,2019)}}                              && $0.51$  & $\bm{0.85}$  && $0.65$  & $\bm{0.83}$ \\
CutMix~\cite{YunHCOYC19} {\small \emph{(eccv,2019)}}                        && $0.51$  & $\bm{0.80}$  && $0.57$  & $\bm{0.82}$ \\
LayerCAM~\cite{JiangZHCW21layercam} {\small \emph{(ieee,2021)}}             && $0.52$  & $\bm{0.86}$  && $0.65$  & $\bm{0.82}$ \\
\cline{1-7} 
\textbf{Transformer-based}  &&  \multicolumn{5}{c|}{} \\
TS-CAM~\cite{gao2021tscam} {\small \emph{(iccv,2021)}}                      && $0.55$  & $\bm{0.88}$  && $0.48$  & $\bm{0.79}$ \\
APViT~\cite{xue2022vision} {\small \emph{(ieee,2022)}}                      && $0.38$  & $\bm{0.85}$  && $0.45$  & $\bm{0.84}$ \\
\cline{1-7}
\end{tabular}
}
\caption{\textbf{Attention-localization (\attcos)} (at layer 5) performance over \rafdb and \affectnet test sets with and without \aus.}
\label{tab:attloc}
\end{table}
}

{
\begin{table}[ht!]
\centering
\resizebox{1\linewidth}{!}{%
\centering
\begin{tabular}{|lllc c c| }
\hline
\multicolumn{3}{|l}{\textbf{Methods}}& &  \cl & \camcos  \\
\hline 
\multirow{1}{*}{CAM~\cite{zhou2016learning} w/o \au }                    && &&$88.20$     &$0.55$\\
\hline 
\multirow{1}{*}{CAM~\cite{zhou2016learning} w/ \au at layers:}           && &&           &      \\
                                                               1          && &&$88.52$     &$0.58$\\
                                                               2          && &&$88.26$     &$0.57$\\
                                                               3          && &&$88.39$     &$0.56$\\
                                                               4          && &&$88.62$     &$0.61$\\
                                                               5          && &&$\bm{88.95}$&$\bm{0.70}$\\
                                                               5+4        && &&$88.78$     &$0.67$\\
                                                               5+4+3      && &&$88.78$     &$0.67$\\
                                                               5+4+3+2    && &&$88.55$     &$0.67$\\
                                                               5+4+3+2+1  && &&$88.46$     &$0.67$\\
\hline
\end{tabular}
}
\caption{Ablation study over \rafdb test set: impact of applying \au alignment over different layers over classification (\cl) and CAM-localization (\camcos). Method: CAM~\cite{zhou2016learning} and ResNet50~\cite{heZRS16} encoder. Layer 1 is the input layer.}
\label{tab:ablation-layer-x}
\end{table}
}

\section{Results and Discussion}
\label{sec:exp}

\subsection{Experimental Methodology}
\label{subsec:results}

\noindent\textbf{Datasets.} To evaluate our method, experiments are conducted on two standard datasets for the FER task: \rafdb~\cite{li17}, and \affectnet~\cite{MollahosseiniHM19}.

\noindent- \emph{\rafdb} Real-world Affective Faces Database is a large scale facial expression dataset~\cite{li17}. Multiple annotators have manually annotated all images. It contains six basic expressions ('Happiness', 'Sadness', 'Surprise', 'Anger', 'Disgust', 'Fear') and 'Neutral'. The dataset contains 12,271 samples for training and 3,068 samples for testing. Both sets have the same distribution.

\noindent- \emph{\affectnet} is one of the largest facial expression recognition datasets with 420k images were manually annotated. Following~\cite{Li21}, 283,901 images are used for training and 3,500 for testing. In terms of labels, this dataset has the same six basic expressions as in the \rafdb dataset, in addition to 'Neutral'.

\noindent\textbf{Implementation Details.} For all our experiments, we trained each method for 100 epochs over \rafdb with a mini-batch size 32 and 20 epochs for \affectnet, with a mini-batch size 224. For CNN-based methods, we used ResNet50~\cite{heZRS16} as an encoder. For transformer-based methods, we used Deit-S~\cite{Touvron21} for TS-CAM~\cite{gao2021tscam}. We used their proposed transformer architecture for the APVIT method~\cite{xue2022vision}, which also relies on Deit-S~\cite{Touvron21}. Images are aligned and resized to ${256\times 256}$, and randomly cropped patches of size ${224\times 224}$ are extracted for training. For APVIT~\cite{xue2022vision}, images are resized to ${112\times 112}$ and fed as input following the method's protocol. The hyper-parameter ${\lambda > 0}$ is estimated using validation by covering values that range from less than 1 to 20. The optimization of training loss is performed using stochastic gradient descent (SGD). The hyper-parameters search of different methods is done over the \rafdb dataset. Due to its large size, relevant hyper-parameters are transferred from \rafdb to \affectnet dataset.

\noindent\textbf{Baseline Methods.} To assess the benefits of using \au cues, we leverage WSOL classifiers. They allow the classification of an image using only the image class as a label. In addition, they provide interpretability maps. In particular, we use CAM-based WSOL methods, which provide interpretability by highlighting ROIs associated with a class. To this end, we selected methods that cover both families of CAM techniques~\cite{rony2023deep}: 1- Bottom-up methods where information flows from the input layer to the output layer. This includes CAM~\cite{zhou2016learning}, WILDCAT~\cite{durand2017wildcat}, ACoL~\cite{ZhangWF0H18}, PRM~\cite{ZhouZYQJ18PRM}, ADL~\cite{ChoeS19}, and TS-CAM~\cite{gao2021tscam}. 2- Top-down methods where information flows in both directions. This includes gradient-based methods: GradCAM~\cite{SelvarajuCDVPB17iccvgradcam}, GradCAM++~\cite{ChattopadhyaySH18wacvgradcampp}, and LayerCAM~\cite{JiangZHCW21layercam}. All these methods are common in the WSOL field. In the FER task, we selected a recent state-of-the-art method, APVIT~\cite{xue2022vision}, which builds spatial attention in its architecture to yield interpretable maps. However, it does not have a CAM module. All the methods are trained with and without our \au loss in addition to classification loss to assess the impact of using \aus as spatial cues for learning. We note that all results are reported using our implementations.

To build \au map cues, we first extract standard 68 facial landmarks from images using SynergyNet~\cite{wu21}\footnote{Other facial landmarks extractors can also easily be used used.}. Following~\cite{Martinez19}, each facial expression can be recognized using only a set of generic \aus. Using the extracted facial landmarks, one can build a 2D map that contains a heatmap of a single \au. For an image, a single 2D heatmap is then built containing all the \aus associated with the image expression. This heatmap encodes the location of the necessary discriminative spatial ROIs needed to recognize the facial expression presented in the image. Training a classifier to focus mainly on these spatial ROIs in an image is expected to improve its interpretability and classification accuracy.
Note that the only manual supervision required for our method is the image expression. No extra manual annotations are required.

\noindent\textbf{Evaluation Metrics.} We use standard classification accuracy (\cl) for the classification task. We also report a classification confusion matrix. Localization of \aus is performed at CAM, and layer-wise attention maps. Although they are both normalized between [0, 1], they do not necessarily have the same value at the pixel level, compared to \au map, since attention is trained to be \emph{correlated} with \au maps. Therefore, we resort to using cosine similarity (Eq.~\ref{eq:cosine}) as a localization (interpretability) measure to assess how a CAM or attention is well aligned with the \au map. To evaluate either a CAM or an attention map, it is first upsampled to the same size as the \au map ${\bm{A}}$. Higher similarity indicates better localization and, hence, better interpretability. We refer to the cosine similarity between a CAM and an \au map as \camcos and as \attcos for the case of attention map versus \au map. Similarly to the classification confusion matrix, we report a per-class cosine similarity matrix. During training, we select the model with the best classification accuracy (\cl), and report its corresponding localization metrics.

\begin{figure}[ht!]
\centering
  \centering
  \includegraphics[width=\linewidth]{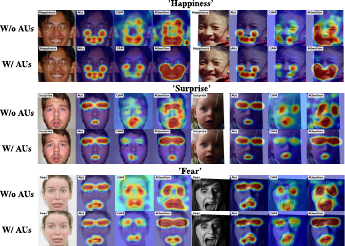}
  \caption[Caption]{Illustration of interpretability prediction over \rafdb test samples using CAM method~\cite{zhou2016learning} with and without action units alignment. From left to right: Input image, true action units map ${\bm{A}}$, CAM ${\bm{M}[k]}$, attention ${\bm{T}_5}$.}
  \label{fig:demo-emo}
\end{figure}

\subsection{Comparison with different methods}
\label{subsec:compare}
\begin{figure}[ht!]
\centering
  \centering
  \includegraphics[width=0.8\linewidth]{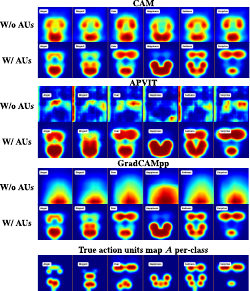}
  \caption[Caption]{Illustration of \textbf{per-class average {\color{red}attention}} maps over all test set of \rafdb with and without action units alignment. Expressions from left to right: 'Anger', 'Disgust', 'Fear', 'Happiness', 'Sadness', 'Surprise'.}
  \label{fig:rafdb-att}
\end{figure}

\begin{figure}[ht!]
\centering
  \centering
  \includegraphics[width=0.8\linewidth]{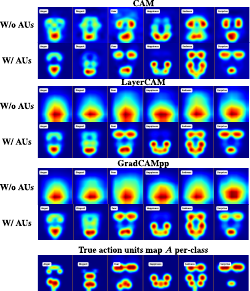}
  \caption[Caption]{Illustration of \textbf{per-class average {\color{red}CAM}} over all test set of \rafdb with and without action units alignment. Expressions from left to right: 'Anger', 'Disgust', 'Fear', 'Happiness', 'Sadness', 'Surprise'.}
  \label{fig:rafdb-cam}
\end{figure}

\begin{figure}[ht!]
\centering
  \centering
  \includegraphics[width=0.8\linewidth]{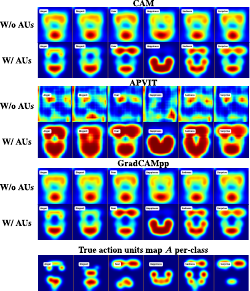}
  \caption[Caption]{Illustration of \textbf{per-class average {\color{red}attention}} maps over all test set of \affectnet with and without action units alignment. Expressions from left to right: 'Anger', 'Disgust', 'Fear', 'Happiness', 'Sadness', 'Surprise'.}
  \label{fig:affectnet-att}
\end{figure}

\begin{figure}[ht!]
\centering
  \centering
  \includegraphics[width=0.8\linewidth]{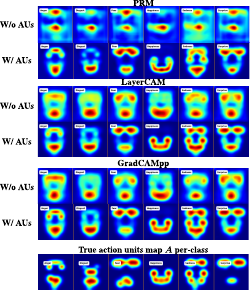}
  \caption[Caption]{Illustration of \textbf{per-class average {\color{red}CAM}} over all test set of \affectnet with and without action units alignment. Expressions from left to right: 'Anger', 'Disgust', 'Fear', 'Happiness', 'Sadness', 'Surprise'.}
  \label{fig:affectnet-cam}
\end{figure}

Table \ref{tab:cl-camloc} shows the impact of using \au alignment over classification and CAM interpretability across different methods. Overall, using our proposed \au alignment helped improve the CAM interpretability by a large margin. Since CAMs are often built after the last layer where we applied the alignment, they are strongly affected, leading to better \au localization. We note that classification accuracy (\cl) has also improved with a relatively large margin over \affectnet compared to \rafdb. This may be related to the fact there is still a large room for improvement in \affectnet compared to \rafdb.
Regarding attention interpretability (Table \ref{tab:attloc}), the effect of using \au loss is larger than in CAM. Most importantly, since the alignment is applied over layer-wise attention, it leads to attention that is better in interpretability than CAM.
Note that achieving better classification accuracy does not necessarily translate to better interpretability. For instance, the vanilla APVIT method~\cite{xue2022vision} has the highest classification accuracy \cl of ${91\%}$ over \rafdb. However, it only scores ${0.38}$ in terms of attention interpretability \attcos, the low est score across all models. Since the vanilla case is optimized only to maximize the classification score, it can easily find features that are not necessarily interpretable to achieve that.
Interestingly, we show that the same model can achieve a similar classification accuracy of ${91.03\%}$ \emph{and} an attention interpretability score of ${0.85}$ when using our alignment loss. This suggests that spatial features have largely shifted their focus, as illustrated in Fig.~\ref{fig:rafdb-att}, without compromising classification performance. As a result, the model finds different spatial features that satisfy both losses: classification and interpretability (\au). 
On \rafdb and \affectnet, we note that our best obtained classification performance of ${91.03\%}$ and ${62.28\%}$ are close to the current state-of-the-art performance of ${92.21\%}$ and ${67.49\%}$ achieved by \cite{Mao23}, respectively.
Additionally, building \au maps during training adds negligible computational time. On the \rafdb dataset, we obtain a training time of 1min 19sec /epoch and 1min 47sec /epoch without and with \au maps. Over the \affectnet dataset, rates of 15min 9sec /epoch and 17min 35sec /epoch are achieved without and with \au maps.
We provide in Fig.\ref{fig:demo-emo} an illustration of prediction with and without our action units alignment. In addition, Figs.~\ref{fig:rafdb-att}, \ref{fig:rafdb-cam}, \ref{fig:affectnet-att}, \ref{fig:affectnet-cam} show per-class average attention and CAM over both dataset. These visuals show better improvement of our classifier interprebility compared to vanilla case.

\begin{figure}[ht!]
\centering
  \centering
  \includegraphics[width=\linewidth]{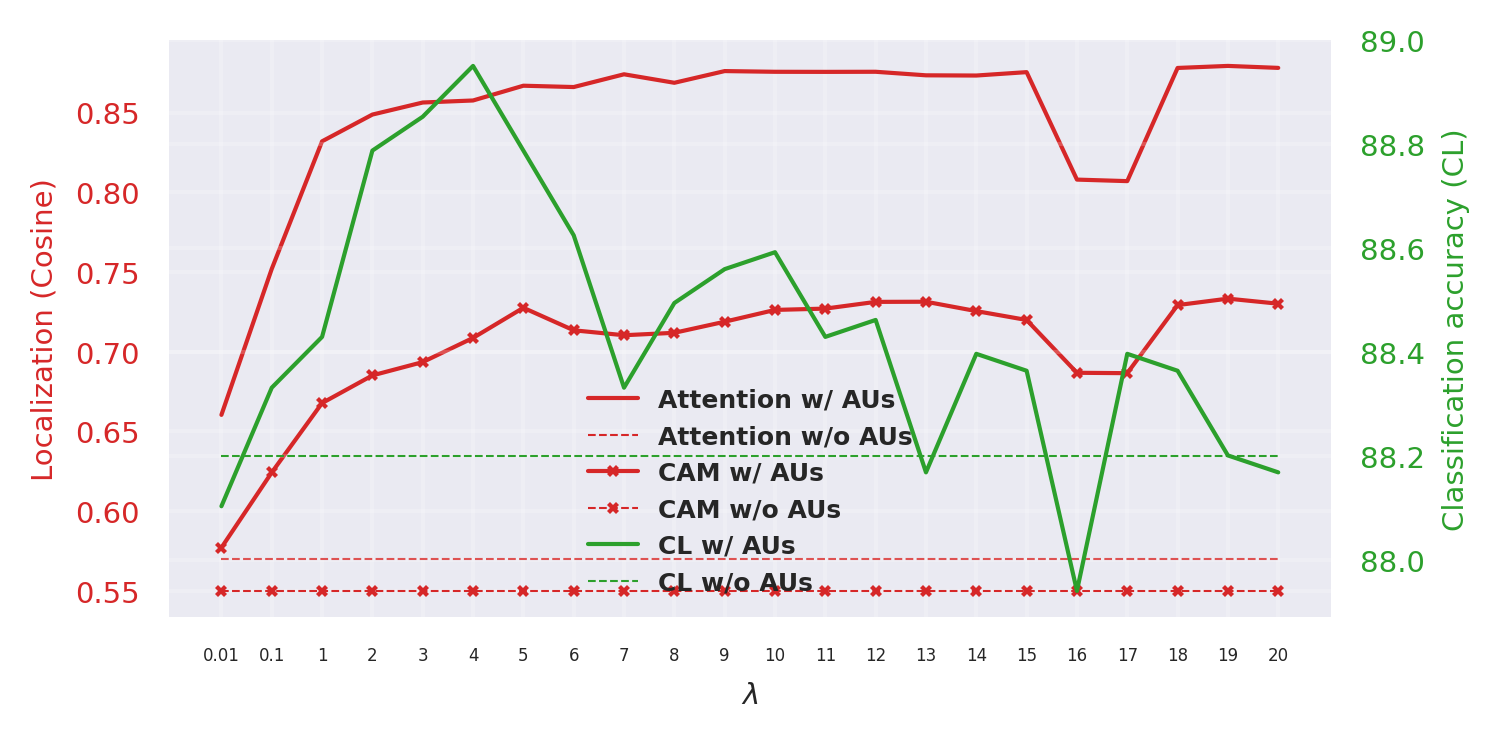}
  \caption[Caption]{Ablations on the \rafdb test set: impact of ${\lambda}$ over classification and localization (interpretability) performance. Alignment to \aus is performed over layer 5 of ResNet50~\cite{heZRS16} with CAM~\cite{zhou2016learning} method.}
  \label{fig:ablation-lambda}
\end{figure}

\subsection{Ablation Studies}
\label{subsec:ablations}
We performed two relevant ablations, assessing how classification and localization (interpretability) performance is affected by ${\lambda}$ and at which layer the alignment loss is applied.

\noindent\textbf{1- Impact of ${\lambda}$} (Fig.~\ref{fig:ablation-lambda}). Given different values of ${\lambda}$, we report in Fig.~\ref{fig:ablation-lambda} its impact on localization (red, left curves) on CAM, and attention at layer 5 using CAM method~\cite{zhou2016learning} and a ResNet50 backbone~\cite{heZRS16} over the \rafdb dataset. It is observed that increasing ${\lambda}$ improves localization at both CAM and attention. However, localization performance starts to stagnate after ${\lambda\approx8}$. Additionally, since the alignment with \aus is performed at the attention level, localization at layer-wise attention (\attcos) is higher than at CAM-level (\camcos). However, they are still better than their corresponding vanilla cases (dashed lines). In practice, although CAMs can be used for interpretability, the attention layer is more accurate and, hence, more reliable.
We also report the impact of ${\lambda}$ on classification accuracy (green, right curve). Increasing ${\lambda}$ showed to improve localization since much importance is considered for the alignment loss term. Classification accuracy also kept improving until ${\lambda\approx4}$. However, large values of ${\lambda}$ put substantial importance on localization compared to classification, leading to degradation in classification accuracy. Hence, a compromise value is the one that improves localization without decreasing classification accuracy (e.g., ${\lambda\approx4}$). Even large ${\lambda}$ values can yield relatively better classification accuracy than the vanilla case (dashed green line).

\noindent\textbf{2- Layer where alignment loss is applied} (Table~\ref{tab:ablation-layer-x}). Overall, applying our \au loss term across single or multiple layers improved both classification and CAM-localization. However, its application on the last feature layer has shown to be most beneficial. Top layers often capture semantic and most discriminative regions~\cite{ZeilerF14}. Since \au maps highlight discriminative ROIs, using them to build better features at top layers is a reasonable choice. In all our experiments, we use the top layer for alignment.



\section{Conclusion}
\label{sec:conlusion}

In this work, we have introduced an approach to make FER models more interpretable. We extended the FER model training with an additional loss that favors feature maps whose average activations are similar to facial \au activations. As in most datasets \au localizations are not available, we estimate them by leveraging off-the-shelf facial point localizers and expert knowledge that associates \aus to the corresponding emotion. By doing that, we manage to obtain a much more reliable and interpretable activation of the FER model that, in many cases, also leads to better recognition performance.

\section*{Acknowledgments}
This work was supported in part by the Fonds de recherche du Québec – Santé (FRQS), the Natural Sciences and Engineering Research Council of Canada (NSERC), Canada Foundation for Innovation (CFI), and the Digital Research Alliance of Canada.

\clearpage
\newpage

\begin{figure*}[ht!]
     \centering
     \begin{subfigure}[b]{0.5\textwidth}
         \centering
         \includegraphics[width=\textwidth]{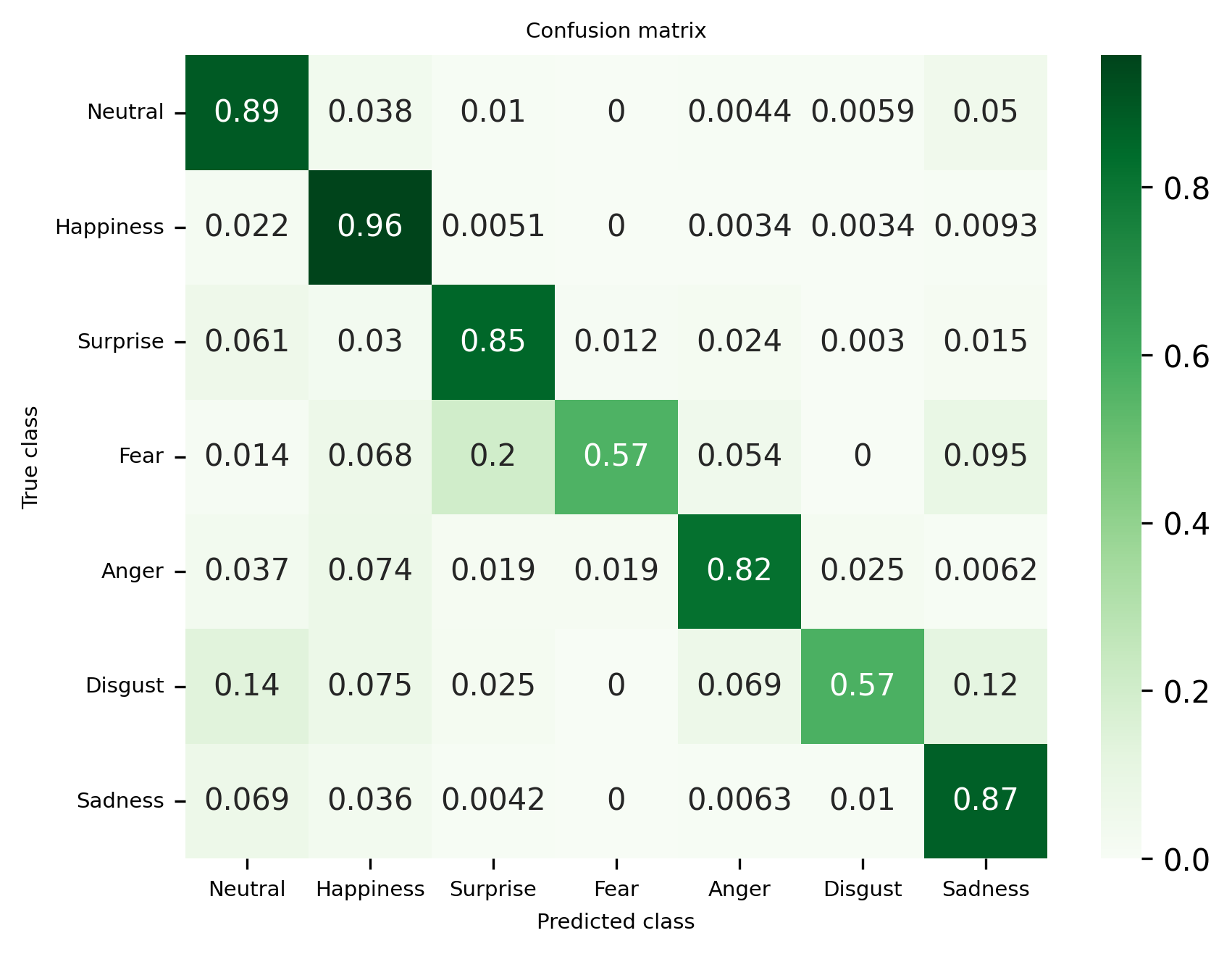}
         \caption{Confusion matrix W/o \aus}
         \label{fig:ablation-range-time}
     \end{subfigure}
     \begin{subfigure}[b]{0.49\textwidth}
         \centering
         \includegraphics[width=\textwidth]{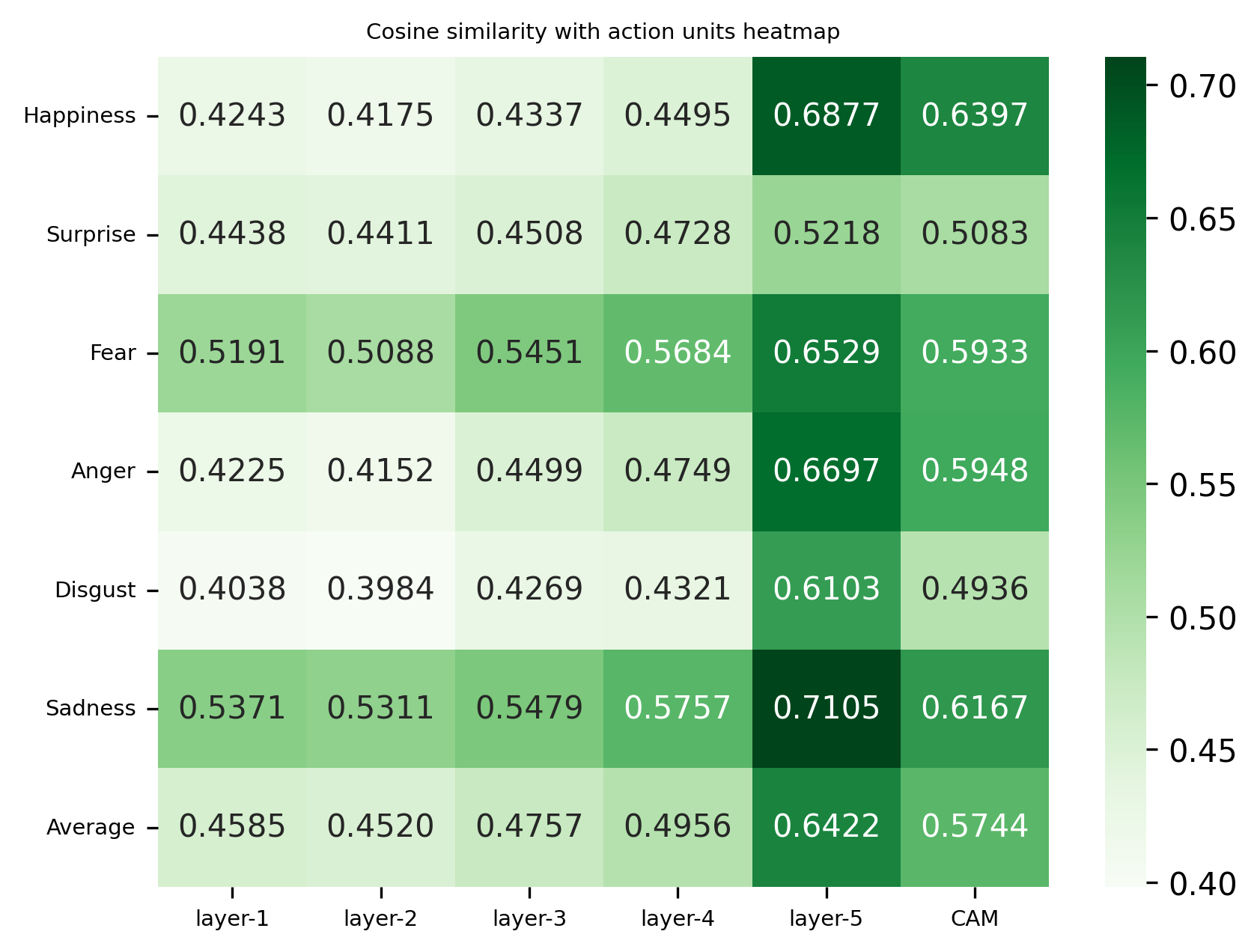}
         \caption{Cosine matrix W/o \aus}
         \label{fig:ablation-lambda-c}
     \end{subfigure}
     \\
     \begin{subfigure}[b]{0.5\textwidth}
         \centering
         \includegraphics[width=\textwidth]{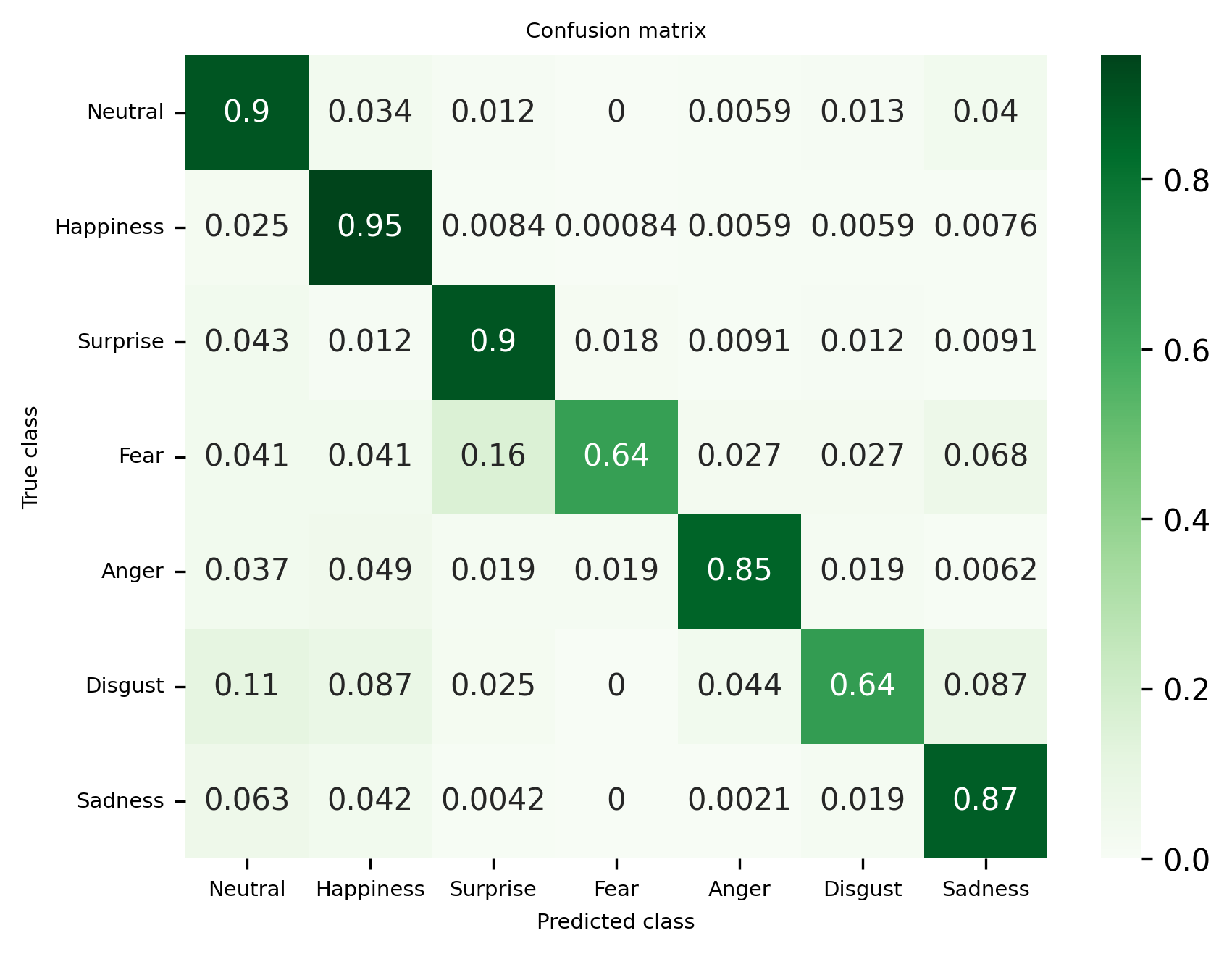}
         \caption{Confusion matrix W/ \aus}
         \label{fig:ablation-lambda-c-const-vs-adap}
     \end{subfigure}
     \begin{subfigure}[b]{0.49\textwidth}
         \centering
         \includegraphics[width=\linewidth]{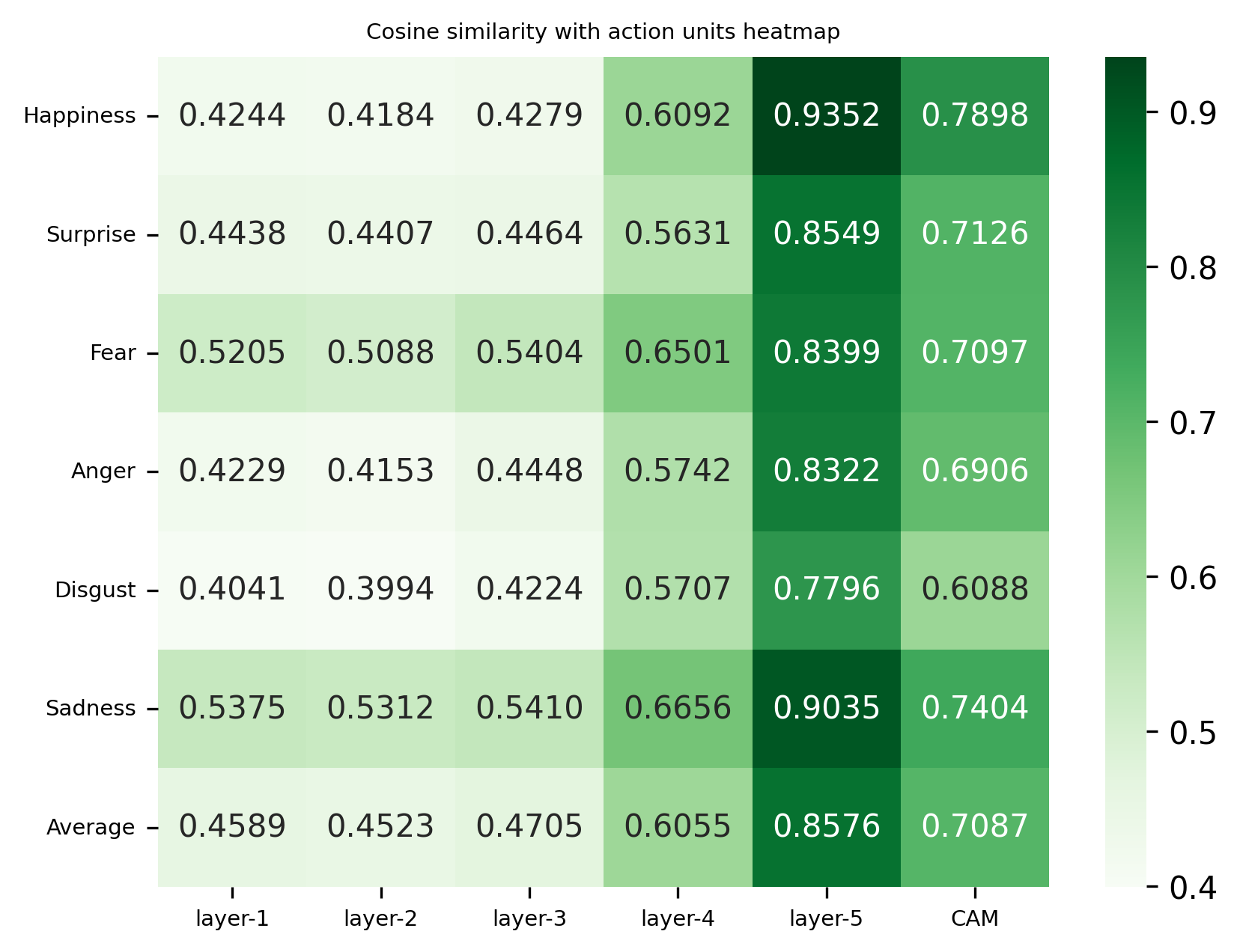}
         \caption{Cosine matrix W/ \aus}
     \label{fig:ablation-time-joint-crf}
     \end{subfigure}
        \caption{Confusion matrix and cosine matrix over \rafdb test set with CAM method~\cite{zhou2016learning} with (top row) and without (bottom row) action units alignment. It show per-class and per-layer/CAM performance.
        }
        \label{fig:per-class-perf-rafdb}
\end{figure*}

\begin{figure*}[ht!]
     \centering
     \begin{subfigure}[b]{0.5\textwidth}
         \centering
         \includegraphics[width=\textwidth]{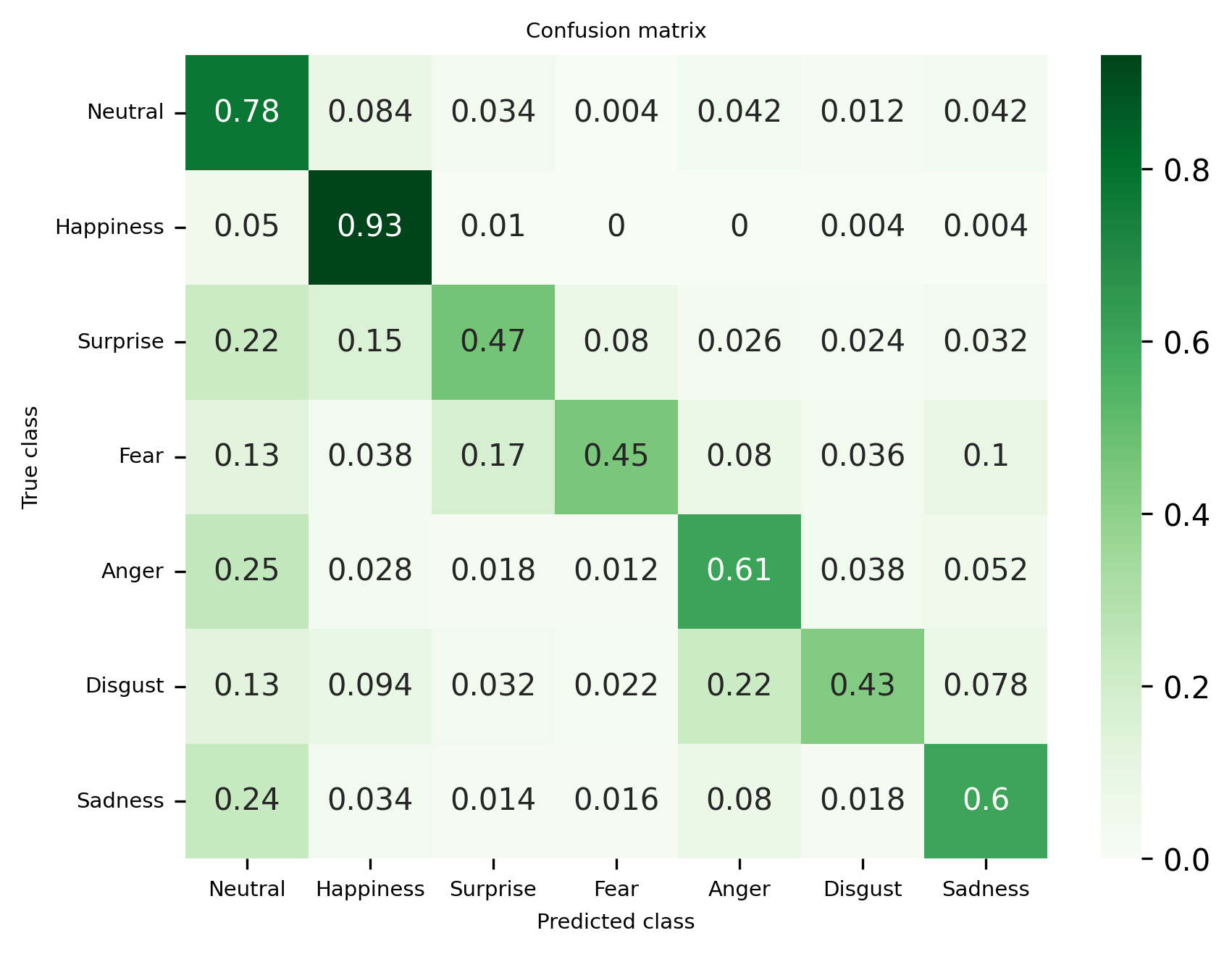}
         \caption{Confusion matrix W/o \aus}
         \label{fig:ablation-range-time}
     \end{subfigure}
     \begin{subfigure}[b]{0.49\textwidth}
         \centering
         \includegraphics[width=\textwidth]{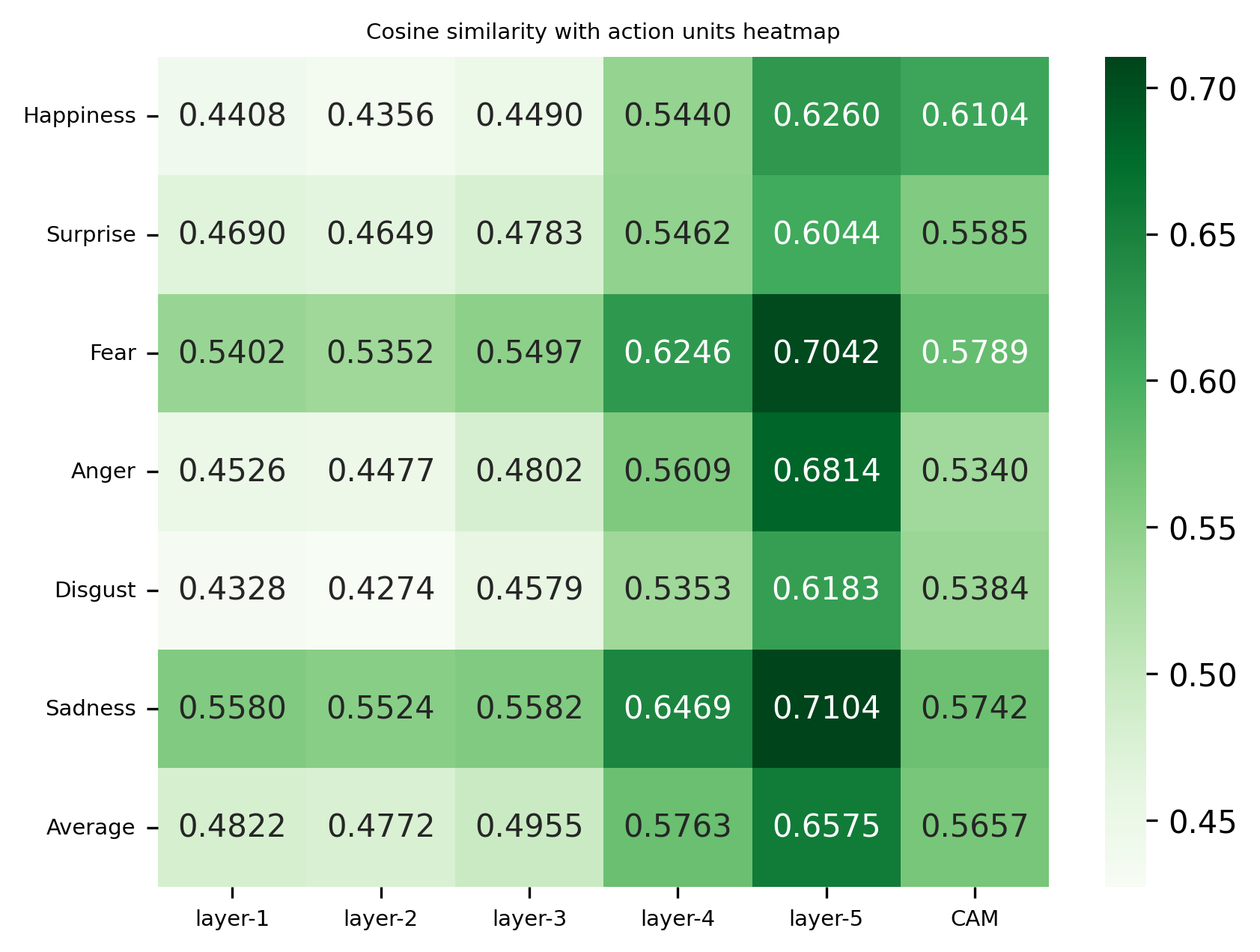}
         \caption{Cosine matrix W/o \aus}
         \label{fig:ablation-lambda-c}
     \end{subfigure}
     \\
     \begin{subfigure}[b]{0.5\textwidth}
         \centering
         \includegraphics[width=\textwidth]{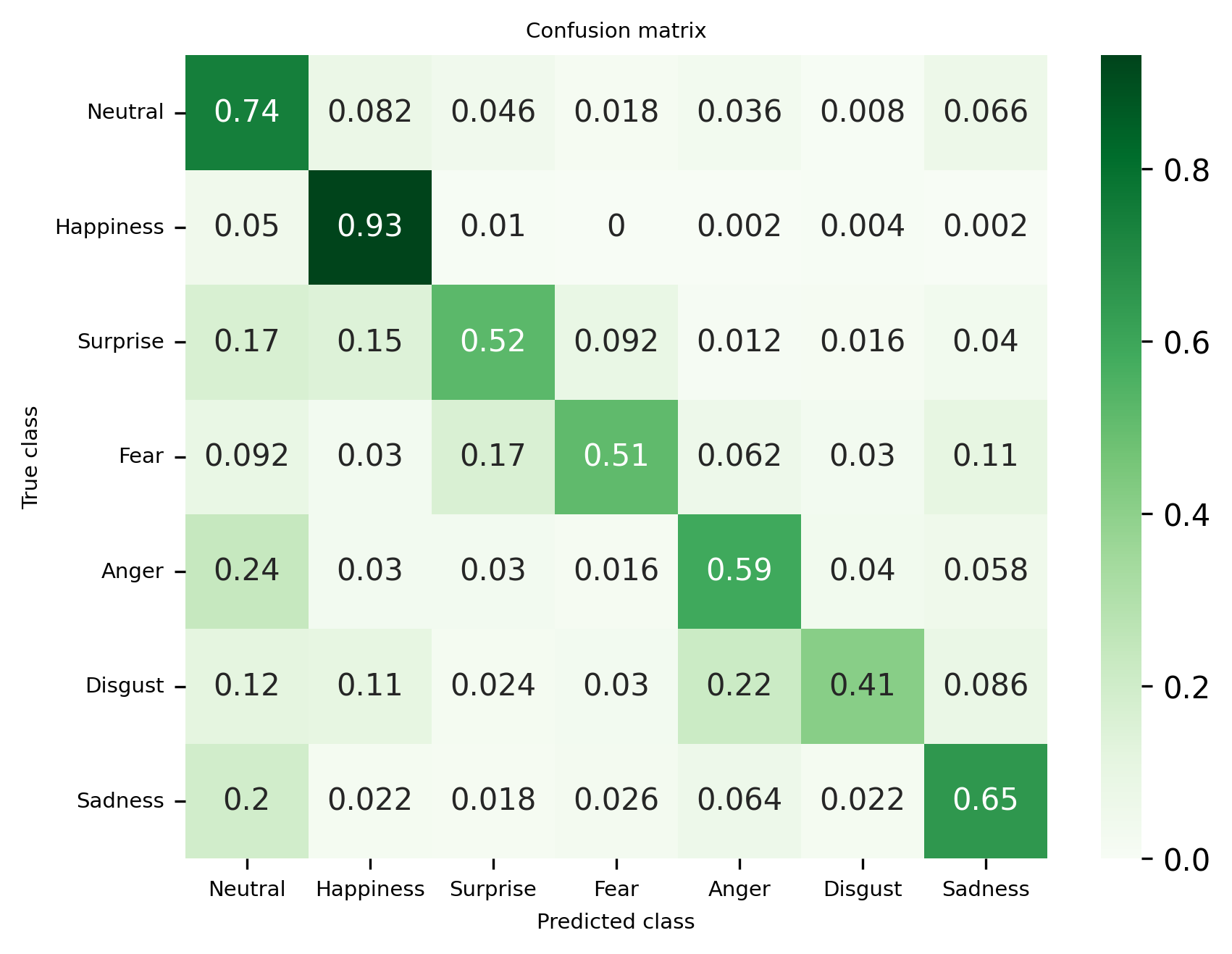}
         \caption{Confusion matrix W/ \aus}
         \label{fig:ablation-lambda-c-const-vs-adap}
     \end{subfigure}
     \begin{subfigure}[b]{0.49\textwidth}
         \centering
         \includegraphics[width=\linewidth]{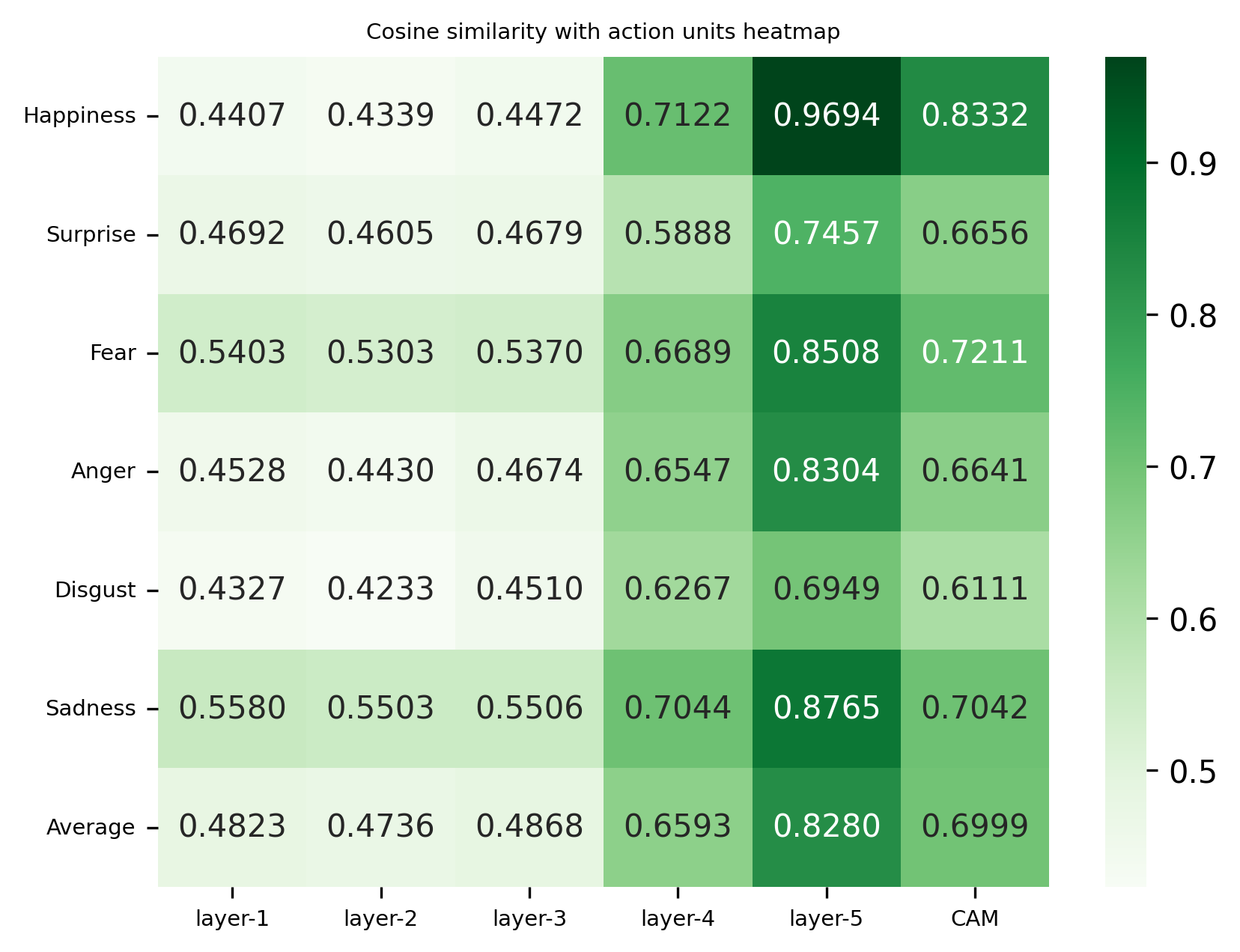}
         \caption{Cosine matrix W/ \aus}
     \label{fig:ablation-time-joint-crf}
     \end{subfigure}
        \caption{Confusion matrix and cosine matrix over \affectnet test set with CAM method~\cite{zhou2016learning} with (top row) and without (bottom row) action units alignment. It show per-class and per-layer/CAM performance.
        }
        \label{fig:per-class-perf-affecnet}
\end{figure*}

\FloatBarrier
\clearpage

\bibliographystyle{abbrv}
\bibliography{biblio}

\end{document}

%% file: macros.tex
\usepackage{times}
\usepackage[numbers]{natbib}
\usepackage[english]{babel}
\usepackage{blindtext}
\usepackage{graphicx}
\usepackage{amsmath, amsthm, amssymb, bbm, bm}
\usepackage{enumerate}
\usepackage{float}      
\usepackage{subcaption}  
\usepackage{wrapfig}
\usepackage[margin=0cm]{caption}
\usepackage[titletoc, toc]{appendix}
\usepackage{tabularx}

\usepackage{multirow}
\usepackage{hhline}
\usepackage{makecell}
\usepackage{placeins}  

\usepackage[x11names, usenames, dvipsnames, svgnames, table]{xcolor}
\definecolor{firebrick}{rgb}{.698,.133,.133}
\definecolor{mybluelight}{rgb}{0.9, 0.9, 1.}
\definecolor{myorangelight}{rgb}{1., 0.9, 0.9}

\usepackage[utf8]{inputenc} 
\usepackage[T1]{fontenc}    
\usepackage{url}            
\usepackage{booktabs, colortbl}       
\usepackage{amsfonts}       
\usepackage{nicefrac}       
\usepackage{microtype}      

\usepackage{csquotes}
\usepackage{latexsym}

\usepackage{pifont}
\usepackage[boxruled, vlined, linesnumbered]{algorithm2e}
\SetAlFnt{\small}
\SetAlCapFnt{\small}
\SetAlCapNameFnt{\small}
\usepackage{algorithmic}
\algsetup{linenosize=\tiny}

\let\oldnl\nl
\newcommand{\nonl}{\renewcommand{\nl}{\let\nl\oldnl}}

\usepackage{paralist}

\usepackage{xspace}
\usepackage{soul}
\usepackage{dsfont}
\usepackage{stmaryrd}
\usepackage[textwidth=15mm]{todonotes}
\usepackage{dirtytalk}
\usepackage{pbox}
\usepackage{cprotect}

\usepackage{verbatim}
\usepackage{textcomp}
\usepackage[normalem]{ulem}

\usepackage{mathtools}
\usepackage{etextools}
\usepackage[inline]{enumitem}

\usepackage[colorlinks=true,allcolors=firebrick,bookmarks=false]{hyperref}

\definecolor{darkergreen}{RGB}{21, 152, 56}
\definecolor{red2}{RGB}{252, 54, 65}
\definecolor{Gray}{gray}{0.85}
\newcolumntype{g}{>{\columncolor{Gray}}c}

\let\OLDthebibliography\thebibliography
\renewcommand\thebibliography[1]{
  \OLDthebibliography{#1}
  \setlength{\parskip}{0pt}
  \setlength{\itemsep}{0pt plus 0.3ex}
}

\newcommand{\reals}{\mathbb{R}}

\theoremstyle{definition}

\DeclarePairedDelimiterX{\divx}[2]{(}{)}{%
  #1\;\delimsize\|\;#2%
}